\documentclass[11pt,a4paper]{article}
\usepackage[hyperref]{acl2019}
\usepackage{times}
\usepackage{latexsym}

\usepackage{url}

\aclfinalcopy 

\usepackage[utf8]{inputenc}
\usepackage{graphicx}
\usepackage{rotating}
\usepackage{multirow}
\usepackage{booktabs}
\usepackage{amsmath}
\usepackage{comment}
\usepackage{xcolor}
\usepackage{colortbl}
\usepackage{amssymb}

\DeclareMathOperator{\nn}{NN}
\DeclareMathOperator{\margin}{margin}

\newcommand{\ds}{\displaystyle}
\newcommand{\MC}{\multicolumn}
\newcommand{\MR}{\multirow}
\newcommand{\ra}{\rightarrow}

\newcommand{\ccnet}{\mbox{CCNet}}
\newcommand{\ccmat}{\mbox{CCMatrix}}
\newcommand{\NbLangs}{38}
\newcommand{\NbLangsTED}{28} 

\newcommand{\NbSents}{32.7}   
\newcommand{\NbSentsWiki}{595}   
\newcommand{\NBMine}{4.5} 
\newcommand{\NbMineEN}{661} 
\newcommand{\NbMineTHIRTYM}{20} 
\newcommand{\NbMineTENM}{112}

\title{CCMatrix: \\Mining Billions of High-Quality Parallel Sentences on the WEB}
\author{
  Holger Schwenk \enskip
  Guillaume Wenzek \enskip
  Sergey Edunov \enskip
  Edouard Grave \enskip
  Armand Joulin \\
  Facebook AI \\
  \texttt{\{schwenk,guw,edunov,egrave,ajoulin\}@fb.com}
  }

\date{}

\begin{document}
\maketitle
\begin{abstract}
We show that margin-based bitext mining in a multilingual sentence space can be applied to monolingual corpora of billions of sentences.
We are using ten snapshots of a curated common crawl corpus \cite{ccnet:2019:arxiv}, totalling \NbSents{} billion unique sentences.
Using one unified approach for \NbLangs{} languages, we were able to mine \NBMine{} billions parallel sentences, out of which \NbMineEN{} million are aligned with English.
\NbMineTHIRTYM{} language pairs have more then 30 million parallel sentences, 
\NbMineTENM{} more then 10 million, and most more than one million, including direct alignments between many European or Asian languages.

To evaluate the quality of the mined bitexts, we train NMT systems for most of the language pairs and evaluate them on TED, WMT and WAT test sets. Using our mined bitexts only and no human translated parallel data, we achieve a new state-of-the-art for a single system on the WMT'19 test set for translation between English and German, Russian and Chinese, as well as German/French. In particular, our English/German system outperforms the best single one by close to 4 BLEU points and is almost on pair with best WMT'19 evaluation system which uses system combination and back-translation.
We also achieve excellent results for distant languages pairs like Russian/Japanese, outperforming the best submission at the 2019 workshop on Asian Translation (WAT).
\end{abstract}

\section{Introduction}
\label{sec:intro}

Most of the current approaches in Natural Language Processing (NLP) are data-driven.
The size of the resources used for training is often the primary concern, but the quality and a large variety of topics may be equally important.
Monolingual texts are usually available in huge amounts for many topics and languages.
However, multilingual resources, typically sentences in two languages which are mutual translations, are more limited, in particular when the two languages do not involve English.
An important source of parallel texts are international organizations like the European Parliament \cite{Koehn:2005:mtsummit_eurparl} or the United Nations \cite{ziemski2016united}. These are professional human translations, but they are in a more formal language and tend to be limited to political topics.
There are several projects relying on volunteers to provide translations for public texts, e.g. news commentary \cite{Tiedmann:2012:lrec_opus}, OpensubTitles \cite{Lison:2016:lrec_opensub} or the TED corpus \cite{Qi:2018:naacl_WordEmbeddings}.

A first system to systematically mine parallel sentences for many language pairs in Wikipedia, including bitexts without English as one of the languages, was presented in \citet{wikimatrix:2019:arxiv}.
In that work, parallel sentence mining was based on a distance measure in a joint multilingual sentence embedding space \cite{Schwenk:2018:acl_mine,Artetxe:2018:mine_arxiv}, using the freely available LASER toolkit\footnote{\url{https://github.com/facebookresearch/LASER}} which provides a language agnostic sentence encoder which was trained on 93 languages \cite{Artetxe:2018:arxiv_massive_ml}.

In this paper, we use the same underlying mining approach based on LASER and scale to a much larger corpus: ten crawls of curated common crawl data set \cite{ccnet:2019:arxiv} instead of Wikipedia (\NbSents{} billion against 550 million unique sentences).
On one hand, we had to redesign the processing pipeline in order to attack the substantial computational challenge: billions of sentence embeddings have to be compared.
One the other hand, it is an interesting research question whether \textbf{global mining} scales to billions of sentences, i.e. systematically comparing each sentence in a source language with all sentences in the target language. 
To the best of our knowledge, all existing large scale bitext mining techniques apply an hierarchical approach. First, a subset of all the texts is selected, e.g. documents, which are supposed to contain parallel sentences. Then, sentences limited to previously aligned documents are compared and the parallel ones are identified.
This type of \textbf{local mining} has the advantage of being very fast since only a few thousand sentences need to be compared for each document. However, sentences which appear in documents which were not preselected can not be aligned.

In this work, we make no assumption on the structure of the monolingual text corpora - we simply compare all sentences against each other. Our experimental results seem to indicate that such an approach works surprisingly well: we are able to mine billions of parallel sentences which seem to be of high quality: NMT systems trained only on our mined data outperform the currently best single NMT systems in WMT'19 and WAT'19.

The paper is organized as follows.
In the next section, we first discuss related work. We then present the corpus used in this work and summarize the underlying mining approach.
Section~\ref{sec:ccmine} describes in detail how we applied this approach to extract parallel sentences.
To asses the quality of the extracted bitexts, we train NMT systems for a subset of language pairs and evaluate them on the TED corpus \cite{Qi:2018:naacl_WordEmbeddings}, test sets of WMT \cite{wmt:2019} and of the the workshop for Asian language (WAT) \cite{wat:2019}. These results are presented in section~\ref{sec:ana_nmt}.
The paper concludes with a discussion of future research directions.

\section{Related work}
\label{sec:related}

There is a large body of research on mining parallel sentences in collections of monolingual texts, usually named \textit{``comparable coprora''}.
Initial approaches to bitext mining have relied on heavily engineered systems often based on metadata information, e.g. \citep{resnik1999mining,resnik2003web}. 
More recent methods explore the textual content of the comparable documents.
For instance, it was proposed to rely on cross-lingual document retrieval, e.g. \citep{utiyama2003reliable,munteanu2005improving} 
or machine translation, e.g. \citep{rauf2009comparable,bouamor2018h2}, typically to obtain an initial alignment that is then further filtered. In the shared task for bilingual document alignment \cite{buck-koehn:2016:WMT1}, many participants used techniques based on $n$-gram or neural language models, neural translation models and bag-of-words lexical translation probabilities  for scoring candidate document pairs. 
The STACC method uses seed lexical translations induced from IBM alignments, which are combined with set expansion operations to score translation candidates through the Jaccard similarity coefficient \citep{etchegoyhen2016set,azpeitia2017weighted,azpeitia2018extracting}.
Using multilingual noisy web-crawls such as ParaCrawl\footnote{\url{http://www.paracrawl.eu/}} for filtering good quality sentence pairs has been explored in the shared tasks for high resource \cite{W18-6453} and low resource \cite{filtering:2019:WMT} languages.

In this work, we rely on massively multilingual sentence embeddings and margin-based mining in the joint embedding space, as described in \cite{Schwenk:2018:acl_mine,Artetxe:2018:mine_arxiv,Artetxe:2018:arxiv_massive_ml}. This approach has also proven to perform best in a low resource scenario \cite{wmt19-filtering-facebook,filtering:2019:WMT}.
Closest to this approach is the research described in \citet{espana2017empirical,hassan2018achieving,Guo:2018:arxiv_mine_bilingual,Yang:2019:arxiv_margin_mine}. However, in all these works, only bilingual sentence representations have been trained. Such an approach does not scale to many languages.
Finally, related ideas have been also proposed in \citet{bouamor2018h2} or \citet{gregoire2017bucc}. However, in those works, mining is not solely based on multilingual sentence embeddings, but they are part of a larger system.

Wikipedia is arguably the largest comparable corpus with high-quality human verified texts. One of the first attempts to exploit this resource was performed by \citet{Adafre:2006:wiki_mine}. An MT system was used to translate Dutch sentences into English and to compare them with the English texts. This method yielded several hundreds of Dutch/English parallel sentences.
Later, a similar technique was applied to the Persian/English pair \cite{Mohammadi:2010:mine_wiki}.
Structural information in Wikipedia such as the topic categories of documents was used in the alignment of multilingual corpora \cite{otero2010wikipedia}.
In another work, the mining approach of \citet{munteanu2005improving} was applied to extract large corpora from Wikipedia in sixteen languages \cite{Smith:2010:naacl_mine_wiki}.
\citet{otero2011measuring} measured the comparability of Wikipedia corpora by the translation equivalents on three languages Portuguese, Spanish, and English.
\citet{patry2011identifying} came up with a set of features such as Wikipedia entities to recognize parallel documents, and their approach was limited to a bilingual setting.
\citet{tufis:2013:ranlp_wiki_mine} proposed an approach to mine parallel sentences from Wikipedia textual content, but they only considered high-resource languages, namely German, Spanish and Romanian paired with English.
\citet{tsai2016cross} grounded multilingual mentions to English wikipedia by training cross-lingual embeddings on twelve languages. 
\citet{gottschalk2017multiwiki} searched for parallel text passages in Wikipedia by comparing their named entities and time expressions.
Finally, \citet{Agha:2018:coling_wiki_mine} propose an approach based on bilingual BiLSTM sentence encoders to mine German, French and Persian parallel texts with English.
Parallel data consisting of aligned Wikipedia titles have been extracted for twenty-three languages.\footnote{\url{https://linguatools.org/tools/corpora/wikipedia-parallel-titles-corpora/}}
Since Wikipedia titles are rarely entire sentences with a subject, verb and object, it seems that only modest improvements were observed when adding this resource to the training material of NMT systems.

We are aware of two large-scale mining approaches applied to several languages pairs and large collections of texts.
The European project ParaCrawl\footnotemark[1] focuses on mining parallel data for all European languages, mainly aligned with English. The underlying alignment engine, called Bitextor,\footnote{\url{https://github.com/bitextor/bitextor}} uses a two stage approach: first parallel documents are identified, and then, pairs of documents are processed to identify parallel segments.
Sentence alignments either uses a seed MT system, or bilingual lexicons \cite{bitextor:2010},
In another work, parallel sentences are mined in Wikipedia for many language pairs using a margin criterion in a multilingual sentence embedding space \cite{wikimatrix:2019:arxiv}


\section{The curated Common Crawl corpus}
\label{sec:ccnet}

\begin{figure}[t]
    \centering
    \includegraphics[width=0.48\textwidth]{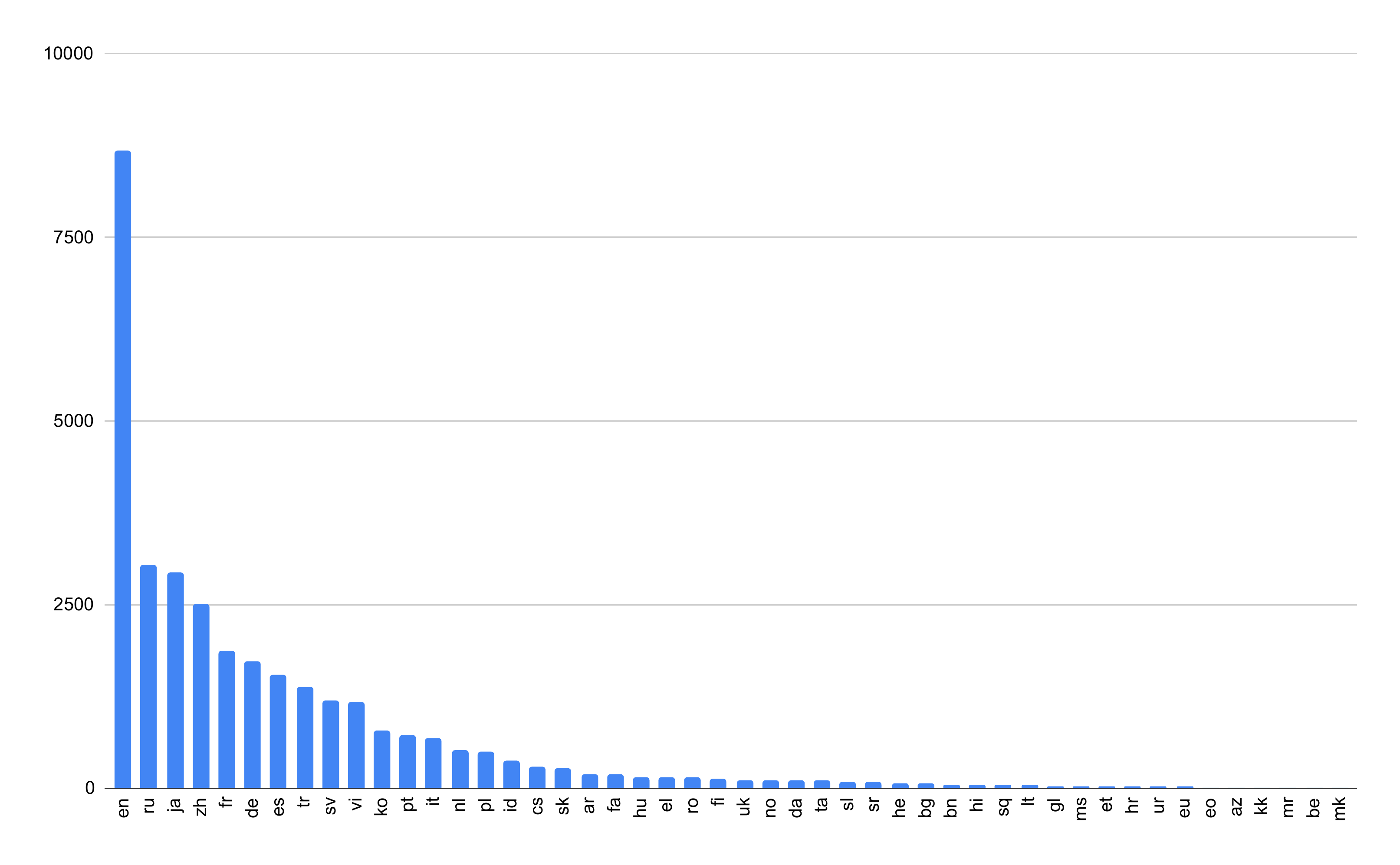}
    \caption{Number of unique sentences in ten crawls of the \ccnet{} corpus.}
    \label{fig:ccnet}
\end{figure}

In this work, we propose to mine parallel sentences from the Web, by using the data released by the Common Crawl project.\footnote{\url{https://commoncrawl.org/}}
Each month, a snapshot of the Web containing terabytes of web pages in various languages is obtained by randomly exploring URLs.
We start by applying some preprocessing steps to the raw text data, following the pipeline introduced by \citet{ccnet:2019:arxiv} and leading to the \ccnet{} dataset.
The first step is to deduplicate the data at the paragraph level, as the original crawls contain up to 70\% of duplicated data.
This preprocessing removes low quality content, such as boilerplate, navigation menus or cookie warnings.
The second step of the pipeline is to identify the language of each document, using fastText\footnote{\url{https://fasttext.cc/docs/en/language-identification.html}} \cite{fasttextlid:2018:arxiv}.
This language identifier uses a linear classifier with character $n$-gram features, and can recognize up to 176 languages.
Finally, the last step of the preprocessing is to filter low quality content by training a language model on Wikipedia, and only keeping documents with a low perplexity score.
We refer the reader to \newcite{ccnet:2019:arxiv} for more details about this preprocessing pipeline.
In Figure~\ref{fig:ccnet}, we report the number of unique sentences obtained after preprocessing ten snapshots from Common Crawl. We currently process \NbLangs{} languages. 
The English Web content is abundant and we used only one snapshot.

\section{Distance-based mining approach}
\label{sec:mine}

\begin{table*}[t!]
  \centering
  \includegraphics[width=0.9\textwidth]{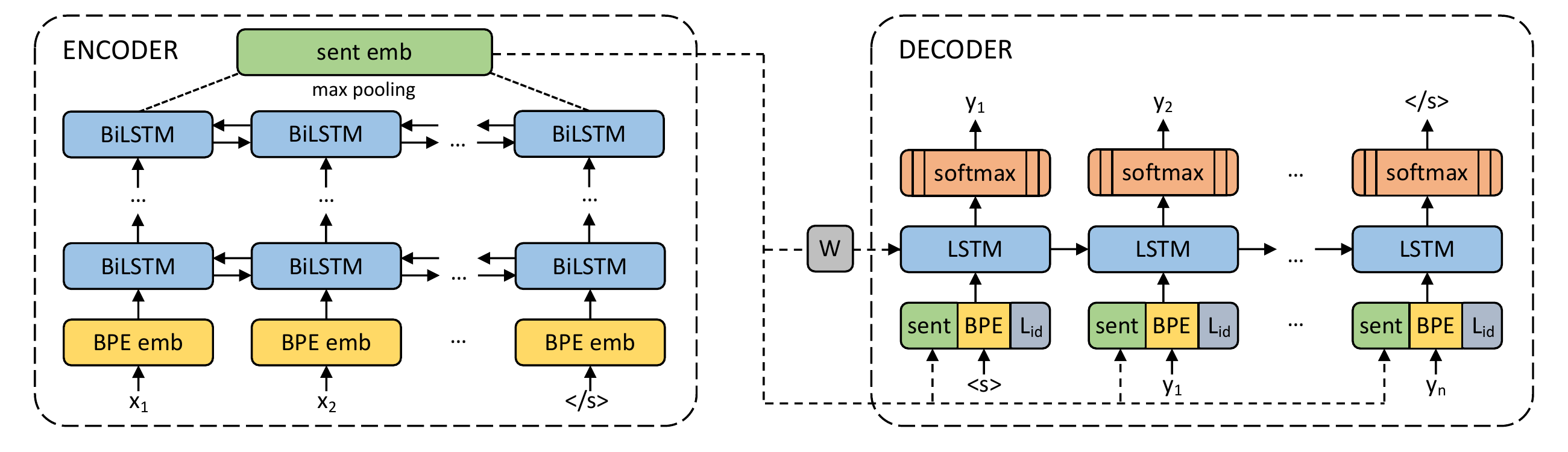}
  \caption{Architecture of the system used to train massively multilingual sentence embeddings.
  See \citet{Artetxe:2018:arxiv_massive_ml} for details.}
  \label{FigLaser}
\end{table*}

The underling idea of the mining approach used in this work is to first learn a multilingual sentence embedding, i.e. an embedding space in which semantically similar sentences are close, independently of the language they are written in. This means that the distance in that space can be used as an indicator whether two sentences are mutual translations or not. Using a simple absolute threshold on the cosine distance was shown to achieve competitive results \cite{Schwenk:2018:acl_mine}.
However, it has been observed that an absolute threshold on the cosine distance is globally not consistent, e.g. \cite{Guo:2018:arxiv_mine_bilingual}.

\subsection{Margin criterion}

\citet{Artetxe:2018:mine_arxiv} showed that the alignment quality can be substantially improved by using a margin criterion instead of an absolute threshold. The margin between two candidate sentences $x$ and $y$ is defined as the ratio between the cosine distance between the two sentence embeddings, and the average cosine similarity of its nearest neighbors in both directions:
\begin{align}
  & \margin(x,y) \nonumber \\
   & \hspace*{4pt}= \frac
   {\ds \cos(x,y)}
   {\ds \sum_{z \in \nn_k(x)}{\hspace*{-8pt}\frac{\cos(x, z)}{2k}} 
     + \hspace*{-6pt}
        \sum_{z \in \nn_k(y)}{\hspace*{-8pt}\frac{\cos(y, z)}{2k}}}
    \label{eqn:margin}        
\end{align}
where $\nn_k(x)$ denotes the $k$ unique nearest neighbors of $x$ in the other language, and analogously for $\nn_k(y)$.

\citet{Artetxe:2018:mine_arxiv} describe the \textit{``max-strategy''} as one of the best performing ones: the margin is first calculated in both directions for all sentences in language $L_1$ and $L_2$. Then, the union of these forward and backward candidates is build, candidates are sorted and pairs with source or target sentences which were already used are omitted. 
Finally, a threshold is applied on the margin score to decide whether two sentences are mutual translations or not.
The reader is referred to \citet{Artetxe:2018:mine_arxiv} for a detailed discussion with related work.
The \textit{``max-strategy''} was used in \citet{wikimatrix:2019:arxiv} to mine parallel sentence in Wikipedia.

This strategy was initially motivated by an evaluation on the BUCC corpus \cite{zweigenbaum2018overview}, for which the reference alignments were known to be strictly \mbox{1:1}.
With increasing corpus size, namely billions of sentences in \ccnet{}, the probability to find several perfect translations increases.
This questions the restriction that each source sentence is aligned to exactly one and only one target sentence, and vice-versa.
The value of $k$ in equation~\ref{eqn:margin} should be also carefully selected to avoid that all the $k$ nearest sentences are valid translations, i.e. having similar distances and therefore a small margin. This would result in many valid translations being excluded.
Therefore, we increased the value of the neighborhood $k$ in Equation~\ref{eqn:margin} from 4, which was used in \cite{wikimatrix:2019:arxiv}, to 16.

\subsection{Multilingual sentence embeddings}
\label{sec:laser}

Distance-based bitext mining requires a joint sentence embedding for all the considered languages. One may be tempted to train a bi-lingual embedding for each language pair, e.g. \cite{espana2017empirical,hassan2018achieving,Guo:2018:arxiv_mine_bilingual,Yang:2019:arxiv_margin_mine}, but this is difficult to scale to thousands of language pairs present in \ccnet{}. We follow \citet{wikimatrix:2019:arxiv} and use one single massively multilingual sentence embedding for all languages, namely the one proposed by the open-source LASER toolkit \cite{Artetxe:2018:arxiv_massive_ml}.

The underlying idea of LASER is to train a sequence-to-sequence system on many language pairs at once using a shared BPE vocabulary and a shared encoder for all languages. The sentence representation is obtained by max-pooling over all encoder output states. Figure~\ref{FigLaser} illustrates this approach.
The reader is referred to \citet{Artetxe:2018:arxiv_massive_ml} for a detailed description.

\begin{figure*}[t!]
    \centering
    \includegraphics[height=5cm]{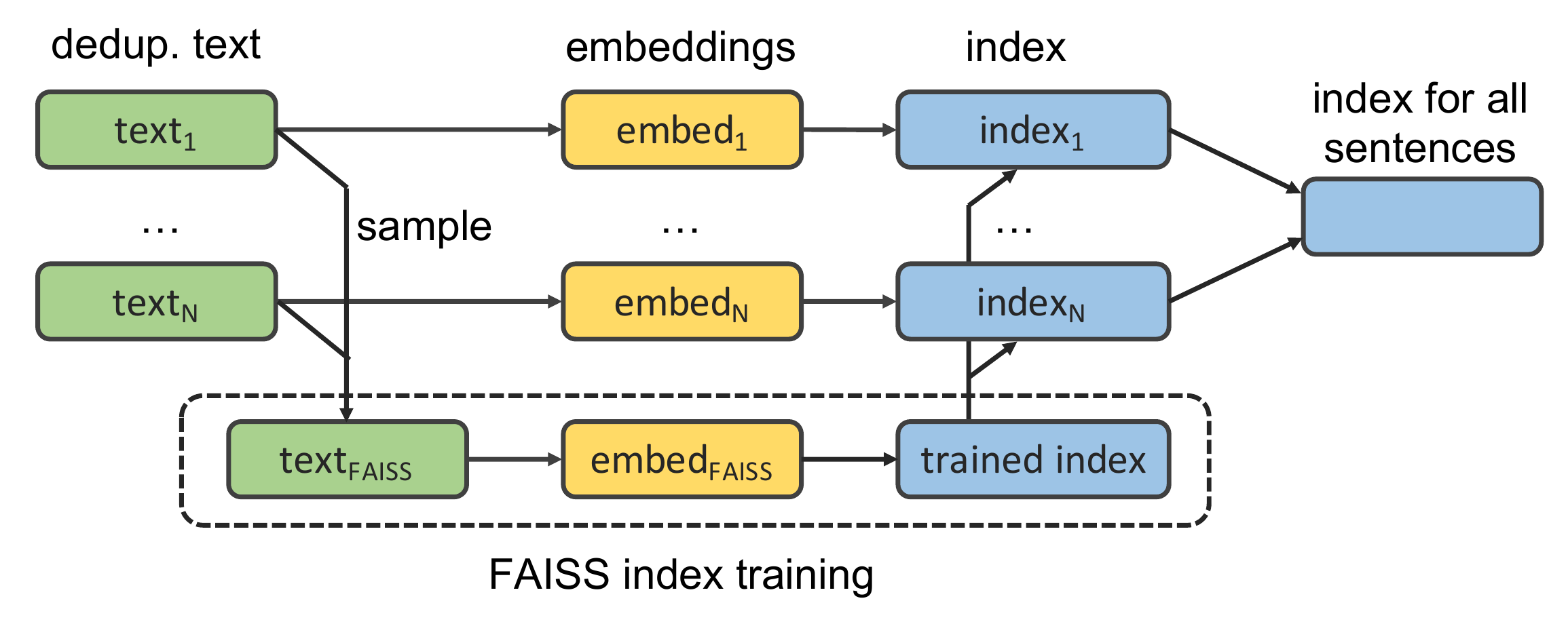}
    \caption{Parallelized processing flow to create an FAISS index for each language.}
   \label{fig:index}
\end{figure*}

\subsection{Scaling to billions of sentences}
\label{sec:faiss}

We use the same underlying mining procedure as \cite{wikimatrix:2019:arxiv} who extracted 135 million parallel sentences from Wikipedia in 1620 different language pairs.
However, our \ccnet{} corpus is more than fifty times larger than Wikipedia: \NbSents{} billion against \NbSentsWiki{} million unique sentences. Our largest corpora are English and Russian, with 8.7 and 3 billion unique sentences, respectively. For ten languages, \ccnet{} has more than one billion unique sentences (see Figure~\ref{fig:ccnet}).
This required to significantly modify the mining pipeline in order to tackle the substantially increased computational complexity. 
The overall processing pipeline can be structured into three tasks:
\begin{enumerate}
    \item text extraction and processing including sentence splitting and language identification;
    \item creation of a compressed index for each language;
    \item mining parallel data for each language pair using the sentence embeddings and indexes.
\end{enumerate}
For each step, we aimed to parallelize the processing as much as possible, by splitting the data into several blocks. We used blocks of about fifty millions sentences. This size was chosen so that the different operations can be performed in a couple of hours. As example, all the English texts are split into 160 blocks.

\subsubsection*{Text extraction}

The first task, text extraction and processing, consists in the following steps:
\begin{itemize}
    \item Extract the texts from the JSON data of \ccnet{} (see \citet{ccnet:2019:arxiv} for details).
    \item Split the \textit{``paragraphs''} into sentences.
    \item Perform LID and exclude sentences which are not in the expected language.
    \item Mark all sentences which are duplicates within each block.
\end{itemize}
Each of these four steps processes are blocks in parallel.
As a final step, we merge all the block-wise deduplicated sentences and create one set of globally unique sentences for each language.
We used a freely available Python tool\footnote{\url{https://pypi.org/project/sentence-splitter/}} to detect sentence boundaries. If specific rules for a language are not available, we fall-back to a linguistically similar languages, e.g. we use Spanish rules for Gallican, and default to English otherwise.
Most of the Asian languages are handled by regular expressions.
We exclude sentences with more than 500 characters.
LID is performed at the sentence level with fastText \cite{joulin:2016:arxiv_ft_lid}.
Once, the text preparation task is finished, we have a corpus of $N_i$ unique sentences for each language $L_i$. These texts are the basis for the index creation and mining tasks.
The amount of data for each language is given in Table~\ref{tab:mine}, third column.

\subsubsection*{Index creation}

\begin{figure*}[t!]
    \centering
    \includegraphics[width=0.9\textwidth]{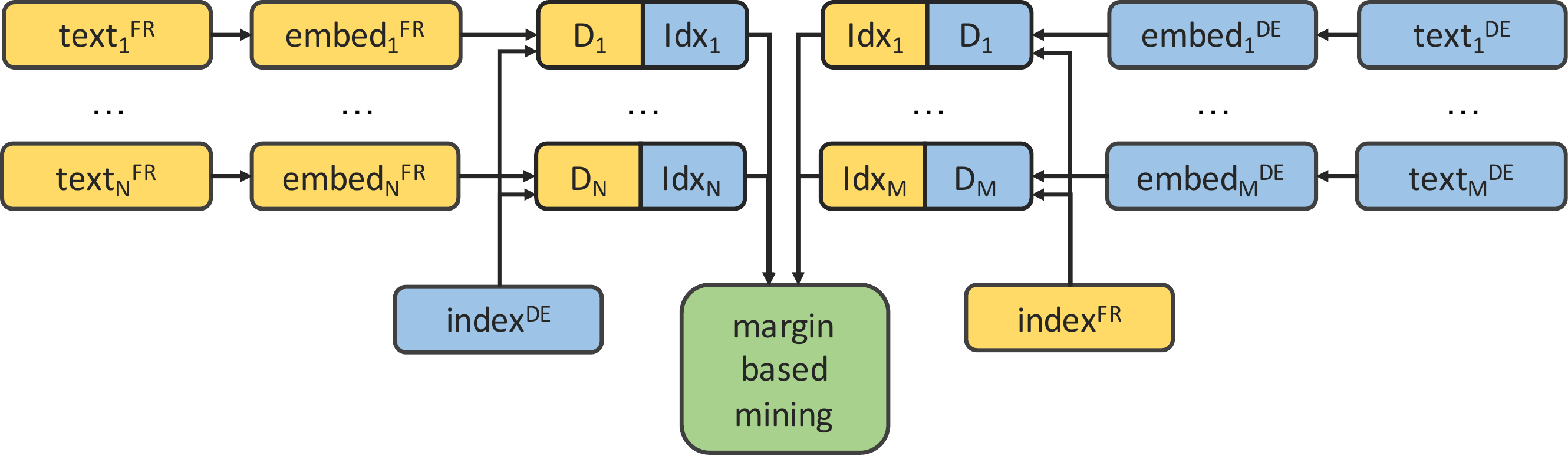}
    \caption{Parallelized processing flow to mine parallel sentences.
        Left: forward distances; Right: backward distances.
        Middle: both distances are combined according to Equation~\ref{eqn:margin}
        and the bitexts extracted.
    }
   \label{fig:mine}
\end{figure*}

We follow \citet{wikimatrix:2019:arxiv} and use the highly optimized FAISS toolkit \cite{FAISS:2017:arxiv}\footnote{\url{https://github.com/facebookresearch/faiss/wiki/Faiss-indexes}} to create compact indexes of the sentence embedding.
LASER's sentence representations are 1024-dimensional. This means that the embeddings of
all sentences would require $\NbSents{}\cdot10^9 \times1024\times4 \approx 130$ TB to store them.
We use an aggressive vector compression based on a 64-bit product-quantizer \cite{Jegou:2011:pami_pq}.
In order to account for the huge number of sentences, we increase the amount of cells from 32k to 64k to partition the search space. This corresponds to
the index type \texttt{OPQ64,IVF65536,PQ64} in FAISS terms.

Exhaustive searching in huge indexes is only tractable if performed on GPU. FAISS supports sharding of a single index on multiple GPUs - this is most efficient if the GPUs are in the same machine and communicate very quickly. For our index type, and eight GPUs with 32GB of memory each, this allows to create an index of about three billion sentences. This includes all languages with the exception of English with 8.7 billion sentences. Therefore, we created three English indexes of 2.7 billion sentences each. 

The processing pipeline to train and create the indexes is summarized in Figure~\ref{fig:index}.
First, we train an index on 40 million sampled sentences of the whole corpus, when available. Once the index is trained, the data in each block is independently added to the common trained index. This can be also processed in parallel. These individual indexes are then merged into one index for each language. The Russian and and Japanese indexes with three billion sentences have a file size of about 200GB, all \NbLangsTED{} indexes total about 2TB.

\subsubsection*{Mining}
\label{sec:ccmine}

Once indexes for all languages are calculated, we can start the mining process for each language pair.
\citet{wikimatrix:2019:arxiv} pre-calculated the sentence embeddings for all languages and then started the pairwise mining process.
The authors report that less than 3.5h on 8 GPUs are needed for the whole \textit{``max-mining''} process between English and German, i.e 134M and 51M sentences respectively. This corresponds to about $1.34\cdot10^8 \times 5.1 \cdot 10^7 \approx 6.8 \cdot 10^{15}$ distances calculations.

Let us consider mining Japanese/Russian bitext in \ccnet{} with 3.0 and 2.9 billion sentences respectively, i.e. $3\cdot10^9 \times 2.9\cdot10^9 \approx 8.7\cdot10^{18}$.
This means that we have to perform about 1300 times more distance calculations, which would translate to more than 6 months on a single machine with 8 GPUs. We tackle this computational challenge by decoupling the distance calculations in forward and backward direction and the margin calculation (see Equation~\ref{eqn:margin}), and processing all these steps in parallel. This processing pipeline is illustrated in Figure~\ref{fig:mine}.

In addition, we had to use a special procedure to mine for parallel sentences with English due to the large amounts of English sentences. For the sake of explanation, let us assume that we want to extract German/English bitexts.
It is computationally too expensive to perform $k$-nn search in the German FAISS index for all the 8.7 billion English sentences (backward distances). Therefore, we are constraint to only use the forward distances $de\ra{}en$. Remember that we had to partition all the English sentences in three indexes of about 2.7 billion sentences each. Consequently, for each German sentence, we search in the three different English indexes, and calculate the margin with respect to the $k=16$ nearest neighbors. We then combine the alignments and keep those which a margin superior to a threshold of 1.06. It can happen that the algorithm finds valid translation in each of the three indexes. We decided to keep those alternative translations.

For all other language pairs $L_1-L_2$, we used the \textit{max-margin} strategy as described in Section~\ref{sec:mine} and Equation~\ref{eqn:margin}, i.e. calculating the forward $L_1\ra{}L_2$ and backward distances $L_2\ra{}L_1$.


\section{Quantitative result analysis}
\label{sec:ana_size}

Mining for parallel sentences in more than 30 billions sentences is computationally very expensive.
In the current version of the \ccmat{} corpus, we have limited the alignment process to \NbLangs{} languages. Those were chosen to cover several language families and scripts.
In the following, we first discuss the amount of extracted sentences. We then turn to a qualitative assessment by training NMT systems for many language pairs (Section~\ref{sec:ana_nmt}).

\subsection{Choosing the margin threshold}
\begin{figure}[b]
    \centering
    \includegraphics{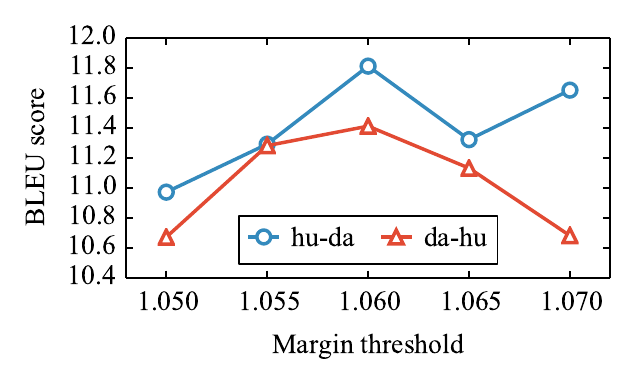}
    \caption{BLEU scores on Hu-Da TED test set for various margin threshold values.}
    \label{fig:margin}
\end{figure}

The margin threshold used to mine parallel sentences will impact the quality of produced bitexts.
A higher threshold will lead to better aligned sentences, and thus higher quality bitexts, but also to smaller datasets.
Thus, there is a trade-off between the size of the extracted bitexts, and their quality.
Exploratory experiments showed that a threshold around $1.06$ seems to give good results.
To confirm this, we trained and evaluated machine translation systems on the Hu-Da pair for different values of the treshold.
We report results in Fig.~\ref{fig:margin}, showing that $1.06$ leads to the best performance.

\subsection{Analysis}

We were able to mine in total \NBMine{} billion parallel sentences when using a threshold of 1.06 on the margin, out of which \NbMineEN{} million are aligned with English (see Table~\ref{tab:mineEn}).

Most of the current MT systems focus on the translation from or into English. Other language pairs are usually handled by pivoting through English since direct parallel texts are much smaller. This can be suboptimal when translating between two morphologically rich languages, e.g. French/German, or very different languages, e.g. Russian/Japanese.
We also provide parallel data for many language pairs not involving English.
Due the high computational complexity, we only considered \NbLangsTED{} languages (see Table~\ref{tab:mine}). This yielded about three million parallel sentence pairs. 
To the best of our knowledge, this makes \ccmat{} the largest collection of high-quality mined parallel texts.
\begin{table}[t!]
  \centering
  \footnotesize
  
\setlength{\tabcolsep}{0.5pt}
\begin{tabular}[t]{lp{5em}p{5em}r@{\hspace{2pt}}r@{\hspace{2pt}}rr}
\toprule
\multirow{2}{*}{\bf ISO} & \multirow{2}{*}{\bf Name} & \multirow{2}{*}{\bf Family} & \multicolumn{2}{c}{\bf\#Sents [M]} & \multicolumn{2}{c}{\bf BLEU} \\
 &&& \textbf{Mono.} & \textbf{Bitext}& \textbf{xx/en} & \textbf{en/xx} \\
 \midrule
\textbf{ar}  & \textbf{Arabic} & \textbf{Arabic} & \textbf{196} & \cellcolor[rgb]{0.812691,0.812691,0.812691} 6.5  & \cellcolor[rgb]{0.641375,0.641375,0.641375} 28.7  & \cellcolor[rgb]{0.792625,0.792625,0.792625} 16.6 \\
\textbf{bg}  & \textbf{Bulgarian} & \textbf{Slavic} & \textbf{68} & \cellcolor[rgb]{0.867983,0.867983,0.867983} 3.7  & \cellcolor[rgb]{0.545875,0.545875,0.545875} 36.3  & \cellcolor[rgb]{0.553625,0.553625,0.553625} 35.7 \\
\textbf{cs}  & \textbf{Czech} & \textbf{Slavic} & \textbf{303} & \cellcolor[rgb]{0.771627,0.771627,0.771627} 9.8  & \cellcolor[rgb]{0.635000,0.635000,0.635000} 29.2  & \cellcolor[rgb]{0.693250,0.693250,0.693250} 24.5 \\
\textbf{da}  & \textbf{Danish} & \textbf{Germanic} & \textbf{109} & \cellcolor[rgb]{0.850271,0.850271,0.850271} 4.5  & \cellcolor[rgb]{0.448750,0.448750,0.448750} 44.1  & \cellcolor[rgb]{0.472625,0.472625,0.472625} 42.2 \\
\textbf{de}  & \textbf{German} & \textbf{Germanic} & \textbf{1728} & \cellcolor[rgb]{0.579151,0.579151,0.579151} 67.3  & \cellcolor[rgb]{0.552125,0.552125,0.552125} 35.8  & \cellcolor[rgb]{0.592125,0.592125,0.592125} 32.6 \\
\textbf{el}  & \textbf{Greek} & \textbf{Hellenic} & \textbf{144} & \cellcolor[rgb]{0.828228,0.828228,0.828228} 5.6  & \cellcolor[rgb]{0.552375,0.552375,0.552375} 35.8  & \cellcolor[rgb]{0.566875,0.566875,0.566875} 34.6 \\
\bottomrule
\end{tabular}

  \caption{CCMatrix: alignments with English. We give the number of the monolingual texts and the extracted parallel sentences (\textbf{all numbers in millions}) for a margin threshold of 1.06, as well as the BLEU scores on the TED test.
  }
  \label{tab:mineEn}
\end{table}

\begin{table*}[t!]
  \centering
  \tiny
  
\setlength{\tabcolsep}{0.5pt}
\begin{tabular}[t]{lp{5em}p{5em}|@{\,\,}r@{\,\,}|@{\,\,}*{28}{r}@{\,\,}|@{\,\,}r}
\toprule
\textbf{ISO} & \textbf{Name} & \textbf{Family} & \textbf{Size}  & \textbf{bg}  & \textbf{cs}  & \textbf{da}  & \textbf{de}  & \textbf{el}  & \textbf{en}  & \textbf{es}  & \textbf{fa}  & \textbf{fi}  & \textbf{fr}  & \textbf{he}  & \textbf{hi}  & \textbf{hu}  & \textbf{id}  & \textbf{it}  & \textbf{ja}  & \textbf{ko}  & \textbf{ms}  & \textbf{nl}  & \textbf{no}  & \textbf{pl}  & \textbf{pt}  & \textbf{ru}  & \textbf{sv}  & \textbf{tr}  & \textbf{uk}  & \textbf{vi}  & \textbf{zh}  & \textbf{Total} \\
\bottomrule
\end{tabular}

  \caption{CCMatrix: number of extracted parallel sentences for each language pair (\textbf{all numbers in millions}) for a margin threshold of 1.06, e.g. we have 33.2 million German/Dutch sentences. The column \textit{``Size''} gives the number of unique sentences (millions) in the monolingual texts after deduplication and LID.
  }
  \label{tab:mine}
\end{table*}
The general tendency is of course that mining in large monolingual corpora leads to larger extracted bitexts. This is however not systematically true.
Let us consider for examples Polish and Dutch which have both about 500 million unique sentences.
When aligned with Czech, a Slavic language, there are slightly more bitexts with Polish than Dutch (13.2M in comparison to 11.6M).
When aligned with German, a Germanic language like Dutch, there are substantially more bitexts for Dutch than Polish, 33.2M and 20.4M respectively.
Finally, both Polish and Dutch have much smaller bitexts with Indonesian although there more than 360M sentences for that language.

One one hand, a possible explanation could be that LASER alignments are more reliable for languages which are very similar, i.e. in the same language family.
On the other hand, it may also be that people which live in nearby countries have similar interests which increases the chance to find translations on the Web.

\section{Qualitative result evaluation}
\label{sec:ana_nmt}

Aiming to perform a large-scale assessment of the quality of the extracted parallel sentences, we trained NMT systems on the extracted parallel sentences and evaluated them on several public test sets.
We identified a publicly available data set which provides test sets for many language pairs: translations of TED talks as proposed in the context of a study on pretrained word embeddings for NMT\footnote{\url{https://github.com/neulab/word-embeddings-for-nmt}} \cite{Qi:2018:naacl_WordEmbeddings}.
The workshop on machine translation (WMT) has a long history of organising evaluations of machine translation, and many comparative results are published for these tasks \cite{wmt:2019}. We provide very competitive BLEU scores for several WMT'19 evaluation tasks in Section~\ref{sec:nmt_wmt}.
Finally, we consider the task of translating between Russian and Japanese as proposed by the 2019 edition of the workshop on Asian translation (see Section~\ref{sec:nmt_wat}).

\subsection{TED corpus}
\label{sec:nmt_ted}

\begin{table*}[t]
  \centering
  \scriptsize
  
\setlength{\tabcolsep}{0.5pt}
\begin{tabular}[t]{lr*{28}{r}r}
\toprule
 & \textbf{ar}  & \textbf{bg}  & \textbf{cs}  & \textbf{da}  & \textbf{de}  & \textbf{el}  & \textbf{en}  & \textbf{es}  & \textbf{fa}  & \textbf{fi}  & \textbf{fr}  & \textbf{he}  & \textbf{hi}  & \textbf{id}  & \textbf{it}  & \textbf{ja}  & \textbf{ko}  & \textbf{ms}  & \textbf{nl}  & \textbf{no}  & \textbf{pl}  & \textbf{pt}  & \textbf{ru}  & \textbf{sv}  & \textbf{tr}  & \textbf{uk}  & \textbf{vi}  & \textbf{zh} \\
\textbf{ar}  & -  & \cellcolor[rgb]{0.793000,0.793000,0.793000} 16.6  & \cellcolor[rgb]{0.856750,0.856750,0.856750} 11.5  & \cellcolor[rgb]{0.813375,0.813375,0.813375} 14.9  & \cellcolor[rgb]{0.805750,0.805750,0.805750} 15.5  & \cellcolor[rgb]{0.780875,0.780875,0.780875} 17.5  & \cellcolor[rgb]{0.641375,0.641375,0.641375} 28.7  & \cellcolor[rgb]{0.720000,0.720000,0.720000} 22.4  & \cellcolor[rgb]{0.891625,0.891625,0.891625} 8.7  & \cellcolor[rgb]{0.908125,0.908125,0.908125} 7.3  & \cellcolor[rgb]{0.754625,0.754625,0.754625} 19.6  & \cellcolor[rgb]{0.863750,0.863750,0.863750} 10.9  & \cellcolor[rgb]{0.842500,0.842500,0.842500} 12.6  & \cellcolor[rgb]{0.780750,0.780750,0.780750} 17.5  & \cellcolor[rgb]{0.767250,0.767250,0.767250} 18.6  & \cellcolor[rgb]{0.894125,0.894125,0.894125} 8.5  & \cellcolor[rgb]{0.965625,0.965625,0.965625} 2.8  & \cellcolor[rgb]{0.872250,0.872250,0.872250} 10.2  & \cellcolor[rgb]{0.802125,0.802125,0.802125} 15.8  & \cellcolor[rgb]{0.797000,0.797000,0.797000} 16.2  & \cellcolor[rgb]{0.872250,0.872250,0.872250} 10.2  & \cellcolor[rgb]{0.741125,0.741125,0.741125} 20.7  & \cellcolor[rgb]{0.827625,0.827625,0.827625} 13.8  & \cellcolor[rgb]{0.764500,0.764500,0.764500} 18.8  & \cellcolor[rgb]{0.905000,0.905000,0.905000} 7.6  & \cellcolor[rgb]{0.910875,0.910875,0.910875} 7.1  & \cellcolor[rgb]{0.762125,0.762125,0.762125} 19.0  & \cellcolor[rgb]{0.876375,0.876375,0.876375} 9.9 \\
\bottomrule
\end{tabular}
  \caption{BLEU scores on the TED test set as proposed in \cite{Qi:2018:naacl_WordEmbeddings}.
  NMT systems were trained on bitexts mined in \ccmat{} only. No other resources were used.
  }
  \label{TabBleuTED}
\end{table*}

In this set of experiments, we are interested in the performance of NMT systems trained on our bitexts only. 
Following~\cite{gottschalk2017multiwiki}, we evaluate on the test sets of the TED dataset \cite{Qi:2018:naacl_WordEmbeddings}. 
This dataset contains parallel TED talk transcripts in 50 languages. 
The TED datasets are tokenized and we first detokenize them using Moses, with the exception of pairs involving the Korean because it creates artifacts. 
As we do not include the training set provided with the TED dataset, we are not guaranteed that our bitexts cover the same domains.

We consider \NbLangsTED{} different languages, resulting in $702$ NMT systems to train.
As a consequence, we apply, if possible, the same pipeline for each pair.
We tokenize the dataset with Moses, with the exception of Chinese where we use Jieba and Japanese where we use Mecab, and compute a BPE vocabulary of size $60$k on the resulting tokenized training bitext. 
Then, for all the pairs, we train the same architecture, that is a Transformer network with $6$ layers for both the encoder and decoder. We use a dimension of $512$ and $4096$ for the feed-forward.
We train each model for $100$ epochs with an initial learning rate of $0.001$.
We keep the model with the best BLEU score on the validation set of TED.

In Table~\ref{TabBleuTED}, we report the BLEU on the test set on the Moses tokenization.
The average BLEU is $16.3$ for all the pairs and $26.9$ for pairs with English.
In comparison with WikiMatrix, we have $46$ pairs out of $702$ with a BLEU above $30$ while they had only $10$ out of $1620$ language pairs.
Their best pair reached $37.3$ BLEU (for Brazilian Portuguese to English), 
while we have $11$ pairs that surpass $37.3$, with our best pair reaching $45.2$ BLEU (Norwegian to English).
These results give an indication on the quality of the mined parallel sentences and suggest that LASER is robust to the noise and difference in domains that exist in a large corpora like Common Crawl.

\subsection{WMT'19 evaluation}
\label{sec:nmt_wmt}

We also evaluate the translation on WMT'19 news translation task. We only consider high resource directions for this comparison as they constitute the biggest challenge, because the existing baseline systems perform strongly and achieving superior performance with mined data is hard. 
We are following the setup described in \citep{ngwmt19} to train systems on En-De, En-Ru, En-Zh and De-Fr. We used Transformer Big architecture with increased FFN size (8192), we trained these models for 500k updates on 8 GPUs with batch size 3500 tokens. Given the large amounts of mined bitexts for the considered language pairs (see Table~\ref{tab:mine}), we limit the sentence pairs to those with score higher than or equal to 1.07 except for En-Zh where we apply filter threshold 1.06. That gives us: 40.6M En-De, 39.5M En-Ru, 32.6M De-Fr and 17.6M En-Zh sentence pairs.
For each direction we learned joined source-target BPE encoding \cite{sennrich2016neural} and used shared input/output embeddings. 
For En-De and En-Ru models we increased model size even further to 9 layers encoder and decoder with layer dropout \cite{fan2019reducing} and increased embed dimensions to 2048. We tuned training parameters on Newstest 2014-2016 when available and on the WMT'19 dev set for De-Fr.  

\begin{table*}[t!]
    \centering
    \begin{tabular}{|@{\,}l@{\,}|l*{8}{c|}}
        \toprule
         \MC{2}{|c}{System} &  de-en & en-de & en-ru & ru-en & zh-en & en-zh & de-fr & fr-de \\
         \midrule
         \MR{4}{*}{\begin{tabular}{l}Single \\ systems\end{tabular}}
         & NT'18 WMT bitext & 46.2 & 45.9 & 33.5 & 33.4 & 25.8 & 39.2 & - & - \\
         & NT'18 \ccmat & \bf 47.4 & \bf 49.7 & \bf 35.4 & \bf 35.3 & 25.8 & \bf 41.3 & - & - \\
         \cmidrule{2-9}
         & NT'19 WMT bitext & 41.0 & 40.4 & 31.4 & 38.1 & - & - & - & - \\
         & NT'19 \ccmat & 40.7 & \bf 44.7 & \bf 34.8 & \bf 39.5 & 29.2 & 34.8 & 37.0 & 33.0 \\
        \midrule
        {\begin{tabular}{l}Ensembles \\ + BT \\ + Reranking \end{tabular}}
         & NT'19 best & 42.8 & 44.9 & 36.3 & 40.1  & 39.3 & 44.6 & 37.3 & 35.0 \\
        \bottomrule
    \end{tabular}
    \caption{BLEU scores on the Newstest'18 (NT'18) and Newstest'19 (NT'19) test set. Newstest'18 WMT  bitext and Newstest'19 WMT bitext are published results for single models trained on parallel WMT'19 data, for En-De and En-Ru results are from \citep{ngwmt19}, for En-Zh results are from \citep{baiduwmt19}. Newstest'19 best are the best BLEU scores achieved by ensembles of models trained on both parallel and back-translated WMT'19 data as of the moment of writing, according to \url{http://matrix.statmt.org/}  }
    \label{tab:nmt_wmt}
\end{table*}

We compare performance of a single model for each direction with the performance of published single models trained on bitext data only. We found that systems trained on \ccmat{} outperform systems trained on bitext data (see Table \ref{tab:nmt_wmt}). This can be seen as a clear indicator of the quality of the mined data.

To answer another question of how does this data combine with real human translated data we train a system using a combination of \ccmat{}{} and bitexts provided by WMT'19, at the example of En-De. We found that this system outperforms the system trained on \ccmat{} data only on average by 0.8 BLEU points achieving BLEU score 50.9 on newstest2018 and 45.1 on newstest2019.

\subsection{WAT'19 evaluation}
\label{sec:nmt_wat}

\begin{table}[b!]
    \centering
    \begin{tabular}{l|l|l}
        \toprule
         System &  Ja / Ru & Ru / Ja \\
         \midrule
         \ccmat{} dev & 16.15 & 19.06 \\
         \ccmat{} test & \bf 14.48 & \bf 18.20 \\
         \midrule
        WAT'19 test best
           & 14.26\footnote{\url{http://lotus.kuee.kyoto-u.ac.jp/WAT/evaluation/list.php?t=67&o=1}}
           & 16.41\footnote{\url{http://lotus.kuee.kyoto-u.ac.jp/WAT/evaluation/list.php?t=66&o=4}} \\
         \bottomrule
    \end{tabular}
    \caption{BLEU scores on the WAT'19 evaluation.}
    \label{tab:nmt_wat}
\end{table}

Finally, we have evaluated the translation between Russian and Japanese as proposed in the 2019 Workshop on Asian Translation (WAT) \cite{wat:2019}.\footnote{http://lotus.kuee.kyoto-u.ac.jp/WAT/WAT2019/index.html}
According to the organizers of the WAT workshop, this language pairs represents \textit{``an extremely low resource situation for distant language pairs''}.
The organizers provide only a tiny amount of parallel data from the Global Voices domain for training (12,356 sentences), and a development (486) and test set (600 sentences) from News Commentary domain, respectively.\footnote{https://github.com/aizhanti/JaRuNC}
The participants in the WAT'19 Russian/Japanese evaluation were encouraged to use other Russian/English and Japanese/English bitexts and train multilingual NMT systems.

We trained an NMT system on \ccmat{} Russian/Japanese bitexts only, without using other resources or texts aligned with English. We applied a threshold of 1.06 on the margin which yielded 9.3 million parallel sentences.
We use the same NMT architecture than in Section~\ref{sec:nmt_wmt}, but with out layer dropout.
We report tokenized BLEU scores using \texttt{multi-bleu.perl} using Moses tokenization for Russian, and Mecab for Japanese (see Table~\ref{tab:nmt_wat}).
We were able to outperform the best performing system at the WAT'19 evaluation (see Table~\ref{tab:nmt_wat}), in particular when translating into Japanese.
The participant in the WAT translation task were constraint to only use the provided resources, which included alignments with English.
Therefore, our results are not directly comparable, but we argue that they are still a good indicator of the alignment quality of our mined bitexts.

\section{Conclusion}

We have shown that margin-based mining in a joint multilingual sentence embedding space can be scaled to monolingual texts of more than 36 billions unique sentences in \NbLangs{} languages.  Our approach is generic and simply compares all sentences among each other, without requiring any document alignment.  We tackled the computational complexity by parallelizing all processing steps.
This procedure yielded \NbMineEN{} million sentences aligned with English, and \NBMine{} billion for pairwise alignments of \NbLangsTED{} languages.  To the best of our knowledge, this is by far the largest collection of high quality parallel sentences.

We have performed an extensive evaluation of the quality of the mined bitexts by training NMT systems for many language pairs.
The mined bitexts seem to be of high quality. Training only on our mined data, we are able to outperform the best reported single NMT system at the WMT'19 evaluations for the translation between German, Russian and Chinese and English, as well as between German and French. We also achieve state-of-the-art BLEU scores for the translation between Russian and Japanese on the WAT'19 test set.
We will provide a script to reproduce our results on the LASER github.\footnote{\url{https://github.com/facebookresearch/LASER}}

In the next version of the \ccmat{} corpus, we will increase the number of common crawl snapshots and focus on low-resource languages. The mined data can be also used to train improved multilingual sentence embeddings.
The large amount of parallel data also raises interesting question how to use it best, for instance, how to efficiently train NMT systems on more than fifty million high quality bitexts?
\section{Acknowledgments}

We would like to thank Matthijs Douze for support with the use of FAISS and Vishrav Chaudhary for helpful comments on this work.

\newpage
\bibliography{CCMatrix}

\begin{thebibliography}{50}
\expandafter\ifx\csname natexlab\endcsname\relax\def\natexlab#1{#1}\fi

\bibitem[{Abdul-Rauf and Schwenk(2009)}]{rauf2009comparable}
Sadaf Abdul-Rauf and Holger Schwenk. 2009.
\newblock \href {http://www.aclweb.org/anthology/E09-1003} {{On the Use of
  Comparable Corpora to Improve SMT performance}}.
\newblock In \emph{EACL}, pages 16--23.

\bibitem[{Adafre and de~Rijke(2006)}]{Adafre:2006:wiki_mine}
Sisay~Fissaha Adafre and Maarten de~Rijke. 2006.
\newblock Finding similar sentences across multiple languages in {W}ikipedia.
\newblock In \emph{Proceedings of the Workshop on {NEW} {TEXT} Wikis and blogs
  and other dynamic text sources}.

\bibitem[{Aghaebrahimian(2018)}]{Agha:2018:coling_wiki_mine}
Ahmad Aghaebrahimian. 2018.
\newblock Deep neural networks at the service of multilingual parallel sentence
  extraction.
\newblock In \emph{Coling}.

\bibitem[{Artetxe and Schwenk(2018{\natexlab{a}})}]{Artetxe:2018:mine_arxiv}
Mikel Artetxe and Holger Schwenk. 2018{\natexlab{a}}.
\newblock {Margin-based Parallel Corpus Mining with Multilingual Sentence
  Embeddings}.
\newblock \emph{\url{https://arxiv.org/abs/1811.01136}}.

\bibitem[{Artetxe and
  Schwenk(2018{\natexlab{b}})}]{Artetxe:2018:arxiv_massive_ml}
Mikel Artetxe and Holger Schwenk. 2018{\natexlab{b}}.
\newblock Massively multilingual sentence embeddings for zero-shot
  cross-lingual transfer and beyond.
\newblock In \emph{\url{https://arxiv.org/abs/1812.10464}}.

\bibitem[{Azpeitia et~al.(2017)Azpeitia, Etchegoyhen, and
  Mart{\'i}nez~Garcia}]{azpeitia2017weighted}
Andoni Azpeitia, Thierry Etchegoyhen, and Eva Mart{\'i}nez~Garcia. 2017.
\newblock \href {http://aclweb.org/anthology/W17-2508} {{Weighted Set-Theoretic
  Alignment of Comparable Sentences}}.
\newblock In \emph{BUCC}, pages 41--45.

\bibitem[{Azpeitia et~al.(2018)Azpeitia, Etchegoyhen, and
  Mart{\'i}nez~Garcia}]{azpeitia2018extracting}
Andoni Azpeitia, Thierry Etchegoyhen, and Eva Mart{\'i}nez~Garcia. 2018.
\newblock {Extracting Parallel Sentences from Comparable Corpora with STACC
  Variants}.
\newblock In \emph{BUCC}.

\bibitem[{Barrault et~al.(2019)Barrault, Bojar, Costa-jussà, Federmann,
  Fishel, Graham, Haddow, Huck, Koehn, Malmasi, Monz, Müller, Pal, Post, and
  Zampieri}]{wmt:2019}
Loïc Barrault, Ondřej Bojar, Marta~R. Costa-jussà, Christian Federmann, Mark
  Fishel, Yvette Graham, Barry Haddow, Matthias Huck, Philipp Koehn, Shervin
  Malmasi, Christof Monz, Mathias Müller, Santanu Pal, Matt Post, and Marcos
  Zampieri. 2019.
\newblock \href {http://www.aclweb.org/anthology/W19-5301} {Findings of the
  2019 conference on machine translation (wmt19)}.
\newblock In \emph{{WMT}}, pages 1--61.

\bibitem[{Bouamor and Sajjad(2018)}]{bouamor2018h2}
Houda Bouamor and Hassan Sajjad. 2018.
\newblock {H2@BUCC18: Parallel Sentence Extraction from Comparable Corpora
  Using Multilingual Sentence Embeddings}.
\newblock In \emph{BUCC}.

\bibitem[{Buck and Koehn(2016)}]{buck-koehn:2016:WMT1}
Christian Buck and Philipp Koehn. 2016.
\newblock \href {http://www.aclweb.org/anthology/W/W16/W16-2347} {Findings of
  the wmt 2016 bilingual document alignment shared task}.
\newblock In \emph{Proceedings of the First Conference on Machine Translation},
  pages 554--563, Berlin, Germany. Association for Computational Linguistics.

\bibitem[{Chaudhary et~al.(2019)Chaudhary, Tang, Guzm\'an, Schwenk, and
  Koehn}]{wmt19-filtering-facebook}
Vishrav Chaudhary, Yuqing Tang, Francisco Guzm\'an, Holger Schwenk, and Philipp
  Koehn. 2019.
\newblock Low-resource corpus filtering using multilingual sentence embeddings.
\newblock In \emph{Proceedings of the Fourth Conference on Machine Translation
  (WMT)}.

\bibitem[{España-Bonet et~al.(2017)España-Bonet, Ádám Csaba~Varga,
  Barrón-Cedeño, and van Genabith}]{espana2017empirical}
Cristina España-Bonet, Ádám Csaba~Varga, Alberto Barrón-Cedeño, and Josef
  van Genabith. 2017.
\newblock {An Empirical Analysis of NMT-Derived Interlingual Embeddings and
  their Use in Parallel Sentence Identification}.
\newblock \emph{IEEE Journal of Selected Topics in Signal Processing}, pages
  1340--1348.

\bibitem[{Esplà-Gomis and Forcada(2010)}]{bitextor:2010}
Miquel Esplà-Gomis and Mikel~L. Forcada. 2010.
\newblock Combining content-based and url-based heuristics to harvest aligned
  bitexts from multilingual sites with bitextor.
\newblock \emph{The Prague Bulletin of Mathematical Linguistics}, 9:77--86.

\bibitem[{Etchegoyhen and Azpeitia(2016)}]{etchegoyhen2016set}
Thierry Etchegoyhen and Andoni Azpeitia. 2016.
\newblock \href {https://doi.org/10.18653/v1/P16-1189} {{Set-Theoretic
  Alignment for Comparable Corpora}}.
\newblock In \emph{ACL}, pages 2009--2018.

\bibitem[{Fan et~al.(2019)Fan, Grave, and Joulin}]{fan2019reducing}
Angela Fan, Edouard Grave, and Armand Joulin. 2019.
\newblock \href {http://arxiv.org/abs/1909.11556} {Reducing transformer depth
  on demand with structured dropout}.

\bibitem[{Gottschalk and Demidova(2017)}]{gottschalk2017multiwiki}
Simon Gottschalk and Elena Demidova. 2017.
\newblock Multiwiki: {I}nterlingual text passage alignment in {W}ikipedia.
\newblock \emph{ACM Transactions on the Web (TWEB)}, 11(1):6.

\bibitem[{Grave et~al.(2018)Grave, Bojanowski, Gupta, Joulin, and
  Mikolov}]{fasttextlid:2018:arxiv}
Edouard Grave, Piotr Bojanowski, Prakhar Gupta, Armand Joulin, and Tomas
  Mikolov. 2018.
\newblock Learning word vectors for 157 languages.
\newblock \emph{\url{https://arxiv.org/abs/1802.06893}}.

\bibitem[{Gr{\'e}goire and Langlais(2017)}]{gregoire2017bucc}
Francis Gr{\'e}goire and Philippe Langlais. 2017.
\newblock \href {http://aclweb.org/anthology/W17-2509} {{BUCC 2017 Shared Task:
  a First Attempt Toward a Deep Learning Framework for Identifying Parallel
  Sentences in Comparable Corpora}}.
\newblock In \emph{BUCC}, pages 46--50.

\bibitem[{Guo et~al.(2018)Guo, Shen, Yang, Ge, Cer, Abrego, Stevens, Constant,
  Sung, Strope, and Kurzweil}]{Guo:2018:arxiv_mine_bilingual}
Mandy Guo, Qinlan Shen, Yinfei Yang, Heming Ge, Daniel Cer, Gustavo~Hernandez
  Abrego, Keith Stevens, Noah Constant, Yun-Hsuan Sung, Brian Strope, and Ray
  Kurzweil. 2018.
\newblock {Effective Parallel Corpus Mining using Bilingual Sentence
  Embeddings}.
\newblock \emph{arXiv:1807.11906}.

\bibitem[{Hassan et~al.(2018)Hassan, Aue, Chen, Chowdhary, Clark, Federmann,
  Huang, Junczys-Dowmunt, Lewis, Li, Liu, Liu, Luo, Menezes, Qin, Seide, Tan,
  Tian, Wu, Wu, Xia, Zhang, Zhang, and Zhou}]{hassan2018achieving}
Hany Hassan, Anthony Aue, Chang Chen, Vishal Chowdhary, Jonathan Clark,
  Christian Federmann, Xuedong Huang, Marcin Junczys-Dowmunt, William Lewis,
  Mu~Li, Shujie Liu, Tie-Yan Liu, Renqian Luo, Arul Menezes, Tao Qin, Frank
  Seide, Xu~Tan, Fei Tian, Lijun Wu, Shuangzhi Wu, Yingce Xia, Dongdong Zhang,
  Zhirui Zhang, and Ming Zhou. 2018.
\newblock {Achieving Human Parity on Automatic Chinese to English News
  Translation}.
\newblock \emph{arXiv:1803.05567}.

\bibitem[{Johnson et~al.(2017)Johnson, Douze, and J{\'e}gou}]{FAISS:2017:arxiv}
Jeff Johnson, Matthijs Douze, and Herv{\'e} J{\'e}gou. 2017.
\newblock Billion-scale similarity search with {GPUs}.
\newblock \emph{arXiv preprint arXiv:1702.08734}.

\bibitem[{Joulin et~al.(2016)Joulin, Grave, Bojanowski, and
  Mikolov}]{joulin:2016:arxiv_ft_lid}
Armand Joulin, Edouard Grave, Piotr Bojanowski, and Tomas Mikolov. 2016.
\newblock Bag of tricks for efficient text classification.
\newblock \emph{\url{https://arxiv.org/abs/1607.01759}}.

\bibitem[{Jégou et~al.(2011)Jégou, Douze, and Schmid}]{Jegou:2011:pami_pq}
H.~Jégou, M.~Douze, and C.~Schmid. 2011.
\newblock Product quantization for nearest neighbor search.
\newblock \emph{IEEE Trans. PAMI}, 33(1):117--128.

\bibitem[{Koehn(2005)}]{Koehn:2005:mtsummit_eurparl}
Philipp Koehn. 2005.
\newblock Europarl: A parallel corpus for statistical machine translation.
\newblock In \emph{MT summit}.

\bibitem[{Koehn et~al.(2019)Koehn, Guzm\'an, Chaudhary, and
  Pino}]{filtering:2019:WMT}
Philipp Koehn, Francisco Guzm\'an, Vishrav Chaudhary, and Juan~M. Pino. 2019.
\newblock Findings of the wmt 2019 shared task on parallel corpus filtering for
  low-resource conditions.
\newblock In \emph{Proceedings of the Fourth Conference on Machine Translation,
  Volume 2: Shared Task Papers}, Florence, Italy. Association for Computational
  Linguistics.

\bibitem[{Koehn et~al.(2018)Koehn, Khayrallah, Heafield, and
  Forcada}]{W18-6453}
Philipp Koehn, Huda Khayrallah, Kenneth Heafield, and Mikel~L. Forcada. 2018.
\newblock \href {https://www.aclweb.org/anthology/W18-6453} {Findings of the
  wmt 2018 shared task on parallel corpus filtering}.
\newblock In \emph{Proceedings of the Third Conference on Machine Translation:
  Shared Task Papers}, pages 726--739, Belgium, Brussels. Association for
  Computational Linguistics.

\bibitem[{Lison and Tiedemann(2016)}]{Lison:2016:lrec_opensub}
P.~Lison and J.~Tiedemann. 2016.
\newblock Opensubtitles2016: Extracting large parallel corpora from movie and
  tv subtitles.
\newblock In \emph{LREC}.

\bibitem[{Mohammadi and GhasemAghaee(2010)}]{Mohammadi:2010:mine_wiki}
Mehdi~Zadeh Mohammadi and Nasser GhasemAghaee. 2010.
\newblock Building bilingual parallel corpora based on {Wikipedia}.
\newblock In \emph{2010 Second International Conference on Computer Engineering
  and Applications}, pages 264--268.

\bibitem[{Munteanu and Marcu(2005)}]{munteanu2005improving}
Dragos~Stefan Munteanu and Daniel Marcu. 2005.
\newblock \href {http://www.aclweb.org/anthology/J05-4003} {{Improving Machine
  Translation Performance by Exploiting Non-Parallel Corpora}}.
\newblock \emph{Computational Linguistics}, 31(4):477--504.

\bibitem[{Nakazawa et~al.(2019)Nakazawa, Doi, Higashiyama, Ding, Dabre, Mino,
  Goto, Pa, Kunchukuttan, Parida, Bojar, and Kurohashi}]{wat:2019}
Toshiaki Nakazawa, Nobushige Doi, Shohei Higashiyama, Chenchen Ding, Raj Dabre,
  Hideya Mino, Isao Goto, Win~Pa Pa, Anoop Kunchukuttan, Shantipriya Parida,
  Ond{\v{r}}ej Bojar, and Sadao Kurohashi. 2019.
\newblock \href {https://doi.org/10.18653/v1/D19-5201} {Overview of the 6th
  workshop on {A}sian translation}.
\newblock In \emph{Proceedings of the 6th Workshop on Asian Translation}, pages
  1--35.

\bibitem[{Ng et~al.(2019)Ng, Yee, Baevski, Ott, Auli, and Edunov}]{ngwmt19}
Nathan Ng, Kyra Yee, Alexei Baevski, Myle Ott, Michael Auli, and Sergey Edunov.
  2019.
\newblock \href {http://www.aclweb.org/anthology/W19-5333} {Facebook fair's
  wmt19 news translation task submission}.
\newblock In \emph{Proceedings of the Fourth Conference on Machine Translation
  (Volume 2: Shared Task Papers, Day 1)}, pages 314--319, Florence, Italy.
  Association for Computational Linguistics.

\bibitem[{Otero et~al.(2011)Otero, L{\'o}pez, Cilenis, and
  de~Compostela}]{otero2011measuring}
P~Otero, I~L{\'o}pez, S~Cilenis, and Santiago de~Compostela. 2011.
\newblock Measuring comparability of multilingual corpora extracted from
  {Wikipedia}.
\newblock \emph{Iberian Cross-Language Natural Language Processings Tasks
  (ICL)}, page~8.

\bibitem[{Otero and L{\'o}pez(2010)}]{otero2010wikipedia}
Pablo~Gamallo Otero and Isaac~Gonz{\'a}lez L{\'o}pez. 2010.
\newblock Wikipedia as multilingual source of comparable corpora.
\newblock In \emph{Proceedings of the 3rd Workshop on Building and Using
  Comparable Corpora, LREC}, pages 21--25.

\bibitem[{Patry and Langlais(2011)}]{patry2011identifying}
Alexandre Patry and Philippe Langlais. 2011.
\newblock Identifying parallel documents from a large bilingual collection of
  texts: Application to parallel article extraction in {Wikipedia}.
\newblock In \emph{Proceedings of the 4th Workshop on Building and Using
  Comparable Corpora: Comparable Corpora and the Web}, pages 87--95.
  Association for Computational Linguistics.

\bibitem[{Qi et~al.(2018)Qi, Sachan, Felix, Padmanabhan, and
  Neubig}]{Qi:2018:naacl_WordEmbeddings}
Ye~Qi, Devendra Sachan, Matthieu Felix, Sarguna Padmanabhan, and Graham Neubig.
  2018.
\newblock \href {http://www.aclweb.org/anthology/N18-2084} {When and why are
  pre-trained word embeddings useful for neural machine translation?}
\newblock In \emph{Proceedings of the 2018 Conference of the North American
  Chapter of the Association for Computational Linguistics: Human Language
  Technologies, Volume 2 (Short Papers)}, pages 529--535, New Orleans,
  Louisiana. Association for Computational Linguistics.

\bibitem[{Resnik(1999)}]{resnik1999mining}
Philip Resnik. 1999.
\newblock \href {http://www.aclweb.org/anthology/P99-1068} {{Mining the Web for
  Bilingual Text}}.
\newblock In \emph{ACL}.

\bibitem[{Resnik and Smith(2003)}]{resnik2003web}
Philip Resnik and Noah~A. Smith. 2003.
\newblock \href {http://www.aclweb.org/anthology/J03-3002} {{The Web as a
  Parallel Corpus}}.
\newblock \emph{Computational Linguistics}, 29(3):349--380.

\bibitem[{Schwenk(2018)}]{Schwenk:2018:acl_mine}
Holger Schwenk. 2018.
\newblock Filtering and mining parallel data in a joint multilingual space.
\newblock In \emph{ACL}, pages 228--234.

\bibitem[{Schwenk et~al.(2019)Schwenk, Chaudhary, Sun, Gong, and
  Guzm{\'{a}}n}]{wikimatrix:2019:arxiv}
Holger Schwenk, Vishrav Chaudhary, Shuo Sun, Hongyu Gong, and Francisco
  Guzm{\'{a}}n. 2019.
\newblock Wikimatrix: Mining 135m parallel sentences in 1620 language pairs
  from wikipedia.
\newblock In \emph{\url{http://arxiv.org/abs/1907.05791}}.

\bibitem[{Sennrich et~al.(2016)Sennrich, Haddow, and
  Birch}]{sennrich2016neural}
Rico Sennrich, Barry Haddow, and Alexandra Birch. 2016.
\newblock Neural machine translation of rare words with subword units.
\newblock In \emph{Proceedings of the 54th Annual Meeting of the Association
  for Computational Linguistics}, pages 1715--1725.

\bibitem[{Smith et~al.(2010)Smith, Quirk, and
  Toutanova}]{Smith:2010:naacl_mine_wiki}
Jason~R. Smith, Chris Quirk, and Kristina Toutanova. 2010.
\newblock Extracting parallel sentences from comparable corpora using document
  level alignment.
\newblock In \emph{NAACL}, pages 403--411.

\bibitem[{Sun et~al.(2019)Sun, Jiang, Xiong, He, Wu, and Wang}]{baiduwmt19}
Meng Sun, Bojian Jiang, Hao Xiong, Zhongjun He, Hua Wu, and Haifeng Wang. 2019.
\newblock \href {http://www.aclweb.org/anthology/W19-5341} {Baidu neural
  machine translation systems for wmt19}.
\newblock In \emph{Proceedings of the Fourth Conference on Machine Translation
  (Volume 2: Shared Task Papers, Day 1)}, pages 374--381, Florence, Italy.
  Association for Computational Linguistics.

\bibitem[{Tiedemann(2012)}]{Tiedmann:2012:lrec_opus}
J.~Tiedemann. 2012.
\newblock Parallel data, tools and interfaces in {OPUS}.
\newblock In \emph{LREC}.

\bibitem[{Tsai and Roth(2016)}]{tsai2016cross}
Chen-Tse Tsai and Dan Roth. 2016.
\newblock Cross-lingual wikification using multilingual embeddings.
\newblock In \emph{Proceedings of the 2016 Conference of the North American
  Chapter of the Association for Computational Linguistics: Human Language
  Technologies}, pages 589--598.

\bibitem[{Tufis et~al.(2013)Tufis, Ion, Ștefan Daniel, Dumitrescu, and
  Ștefănescu}]{tufis:2013:ranlp_wiki_mine}
Dan Tufis, Radu Ion, Ștefan Daniel, Dumitrescu, and Dan Ștefănescu. 2013.
\newblock Wikipedia as an smt training corpus.
\newblock In \emph{RANLP}, pages 702--709.

\bibitem[{Utiyama and Isahara(2003)}]{utiyama2003reliable}
Masao Utiyama and Hitoshi Isahara. 2003.
\newblock \href {http://www.aclweb.org/anthology/P03-1010} {{Reliable Measures
  for Aligning Japanese-English News Articles and Sentences}}.
\newblock In \emph{ACL}.

\bibitem[{Wenzek et~al.(2019)Wenzek, Lachaux, Conneau, Chaudhary, Guzm\'an,
  Joulin, and Grave}]{ccnet:2019:arxiv}
Guillaume Wenzek, Marie-Anne Lachaux, Alexis Conneau, Vishrav Chaudhary,
  Francisco Guzm\'an, Armand Joulin, and Edouard Grave. 2019.
\newblock {CCNet:} extracting high quality monolingual datasets from web crawl
  data.
\newblock \emph{\url{https://arxiv.org/abs/1911.00359}}.

\bibitem[{Yang et~al.(2019)Yang, Ábrego, Yuan, Guo, Shen, Cer, Sung, Strope,
  and Kurzweil}]{Yang:2019:arxiv_margin_mine}
Yinfei Yang, Gustavo~Hernández Ábrego, Steve Yuan, Mandy Guo, Qinlan Shen,
  Daniel Cer, Yun-Hsuan Sung, Brian Strope, and Ray Kurzweil. 2019.
\newblock Improving multilingual sentence embedding using bi-directional dual
  encoder with additive margin softmax.
\newblock In \emph{\url{https://arxiv.org/abs/1902.08564}}.

\bibitem[{Ziemski et~al.(2016)Ziemski, Junczys-Dowmunt, and
  Pouliquen}]{ziemski2016united}
Michał Ziemski, Marcin Junczys-Dowmunt, and Bruno Pouliquen. 2016.
\newblock {The United Nations Parallel Corpus v1.0}.
\newblock In \emph{LREC}.

\bibitem[{Zweigenbaum et~al.(2018)Zweigenbaum, Sharoff, and
  Rapp}]{zweigenbaum2018overview}
Pierre Zweigenbaum, Serge Sharoff, and Reinhard Rapp. 2018.
\newblock \href {http://lrec-conf.org/workshops/lrec2018/W8/pdf/12_W8.pdf}
  {{Overview of the Third BUCC Shared Task: Spotting Parallel Sentences in
  Comparable Corpora}}.
\newblock In \emph{Proceedings of the 11th Workshop on Building and Using
  Comparable Corpora}.

\end{thebibliography}
\bibliographystyle{acl_natbib}

\end{document}